%% file: full.tex
\documentclass{article}

\usepackage{microtype}
\usepackage{graphicx}
\usepackage{subcaption}
\usepackage{booktabs} 

\usepackage{hyperref}

\usepackage{amsmath}
\DeclareMathOperator*{\argmax}{arg\,max}

\DeclareMathOperator*{\argtop}{arg\,top}
\usepackage{amsthm}
\newtheorem{theorem}{Theorem}

\theoremstyle{definition}

\usepackage{amssymb}
\usepackage{bm}
\usepackage{bbm}
\usepackage{url}
\usepackage{todonotes}
\newcommand{\todoherke}[1]{\todo[color=green!40]{#1}}
\renewcommand{\todoherke}[1]{}
\renewcommand{\todo}[1]{}
\newcommand{\bigO}{\mathcal{O}}
\DeclareMathOperator{\logonemexp}{log1mexp}
\DeclareMathOperator{\logonepexp}{log1pexp}
\DeclareMathOperator{\expmone}{expm1}
\DeclareMathOperator{\logonep}{log1p}

\let\emptyset\varnothing
\newcommand\itemmargin{-0.2em}

\usepackage{tikz}
\usetikzlibrary{decorations.pathreplacing}

\usepackage[accepted]{icml2019}

\icmltitlerunning{Stochastic Beams and Where to Find Them}

\begin{document}

\include{main}

\include{appendix}

\end{document}

%% file: main.tex
\twocolumn[
\icmltitle{Stochastic Beams and Where to Find Them: \\ The Gumbel-Top-$k$ Trick for Sampling Sequences Without Replacement}



\icmlsetsymbol{equal}{*}

\begin{icmlauthorlist}
\icmlauthor{Wouter Kool}{uva,ortec}
\icmlauthor{Herke van Hoof}{uva}
\icmlauthor{Max Welling}{uva,cifar}
\end{icmlauthorlist}

\icmlaffiliation{uva}{University of Amsterdam, The Netherlands}
\icmlaffiliation{ortec}{ORTEC, The Netherlands}
\icmlaffiliation{cifar}{CIFAR, Canada}

\icmlcorrespondingauthor{Wouter Kool}{w.w.m.kool@uva.nl}

\icmlkeywords{Machine Learning, Sequence Modelling, Stochastic Beam Search, Gumbel-Max trick, Gumbel-Top-k trick, Sampling without replacement}

\vskip 0.3in
]



\printAffiliationsAndNotice{} 

\begin{abstract}
The well-known Gumbel-Max trick for sampling from a categorical distribution can be extended to sample $k$ elements without replacement. We show how to implicitly apply this `\emph{Gumbel-Top-$k$}' trick on a factorized distribution over sequences, allowing to draw exact samples without replacement using a \emph{Stochastic Beam Search}. Even for exponentially large domains, the number of model evaluations grows only linear in $k$ and the maximum sampled sequence length. The algorithm creates a theoretical connection between sampling and (deterministic) beam search and can be used as a principled intermediate alternative. In a translation task, the proposed method compares favourably against alternatives to obtain diverse yet good quality translations. We show that sequences sampled without replacement can be used to construct low-variance estimators for expected sentence-level BLEU score and model entropy.
\end{abstract}

\section{Introduction}
We think the Gumbel-Max trick \cite{gumbel1954statistical,maddison2014sampling} is like a \emph{magic} trick. It allows sampling from the categorical distribution, simply by \emph{perturbing} the log-probability for each category by adding \emph{independent} Gumbel distributed noise and returning the category with maximum \emph{perturbed log-probability}. This trick has recently (re-)gained popularity as it allows to derive reparameterizable continous relaxations of the categorical distribution \citep{maddison2016concrete,jang2016categorical}. However, there is more: as was noted (in a blog post) by \citet{vieira2014gumbel}, taking the top $k$ largest perturbed log-probabilities (instead of the maximum, or \emph{top 1}) yields a sample of size $k$ from the categorical distribution \emph{without replacement}. We refer to this extension of the Gumbel-Max trick as the \emph{Gumbel-Top-$k$} trick.

In this paper, we consider \emph{factorized distributions over sequences}, represented by (parametric) \emph{sequence models}. Sequence models are widely used, e.g. in tasks such as neural machine translation \citep{sutskever2014sequence, bahdanau2014neural} and image captioning \citep{vinyals2015show}. Many such tasks require obtaining a set of representative sequences from the model. These can be random samples, but for low-entropy distributions, a set of sequences sampled using standard sampling (with replacement) may contain many duplicates. On the other hand, a \emph{beam search} can find a set of unique high-probability sequences, but these have low variability and, being deterministic, cannot be used to construct statistical estimators.

In this paper, we propose sampling without replacement as an alternative method to obtain representative sequences from a sequence model. We show that for a sequence model, we can draw samples without replacement by applying the Gumbel-Top-$k$ trick \emph{implicitly}, without instantiating all sequences in the (typically exponentially large) domain. This procedure allows us to draw unique sequences using a number of model evaluations that grows only \emph{linear} in the number of samples $k$ and the maximum sampled sequence length. The algorithm uses the idea of top-down sampling \citep{maddison2014sampling} and performs a beam search over stochastically perturbed log-probabilities, which is why we refer to it as \emph{Stochastic Beam Search}. 

\textit{Stochastic Beam Search is a novel procedure for sampling sequences that avoids duplicate samples.} Unlike ordinary beam search, \textit{it has a probabilistic interpretation.} Thus, it can e.g. be used in importance sampling. As such, Stochastic Beam Search conceptually connects sampling and beam search, and combines advantages of both methods. In Section \ref{sec:experiments} we give two examples of how Stochastic Beam Search can be used as a principled alternative to sampling or beam search. In these experiments, Stochastic Beam Search is used to control the diversity of translation results, as well as to construct low variance estimators for sentence-level BLEU score and model entropy.

\section{Preliminaries}
\label{sec:preliminaries}
\subsection{The categorical distribution}
A discrete random variable $I$ has a distribution $\text{Categorical}(p_1, ..., p_n)$ with domain $N = \{1, ..., n\}$ if $P(I=i) = p_i \quad \forall i \in N$.
We refer to the categories $i$ as \emph{elements} in the domain $N$ and denote with $\phi_i, i \in N$ the \emph{(unnormalized) log-probabilities}, so $\exp{\phi_i} \propto p_i$ and $p_i = \frac{\exp{\phi_i}}{\sum_{j \in N} \exp{\phi_j}}$. Therefore in general we write:
\begin{equation}
\label{eq:categorical}
    I \sim \text{Categorical} \left( \frac{\exp{\phi_i}}{\sum_{j \in N} \exp{\phi_j}}, i \in N \right).
\end{equation}

\subsection{The Gumbel distribution}
\label{sec:gumbel_distribution}
If $U \sim \text{Uniform}(0,1)$, then $G = \phi - \log (- \log U)$ has a \emph{Gumbel} distribution with location $\phi$ and we write $G \sim \text{Gumbel}(\phi)$. From this it follows that $G' = G + \phi' \sim \text{Gumbel}(\phi + \phi')$, so we can \emph{shift} Gumbel variables.

\subsection{The Gumbel-Max trick}
\label{sec:gumbel_max_trick}
The Gumbel-Max trick \citep{gumbel1954statistical, maddison2014sampling} allows to sample from the categorical distribution \eqref{eq:categorical} by independently \emph{perturbing} the log-probabilities $\phi_i$ with Gumbel noise and finding the largest element.

Formally, let $G_i \sim \text{Gumbel}(0), i \in N$ i.i.d. and let $I^* = \argmax_i \{ \phi_i + G_i \}$, then $I^* \sim \text{Categorical}(p_i, i \in N)$ with $p_i \propto \exp \phi_i$.
In a slight abuse of notation, we write $G_{\phi_i} = G_i + \phi_i \sim \text{Gumbel}(\phi_i)$ and we call $G_{\phi_i}$ the \emph{perturbed log-probability} of element or category $i$ in the domain $N$.

For any subset $B \subseteq N$ it holds that \citep{maddison2014sampling}:
\begin{align}
\label{eq:gumbel_max}
    \max_{i \in B} G_{\phi_i} & \, \sim \text{Gumbel} \left( \log \sum_{j \in B} \exp{\phi_j}\right), \\
\label{eq:gumbel_argmax}
    \argmax_{i \in B} G_{\phi_i} & \, \sim \text{Categorical} \left( \frac{\exp{\phi_i}}{\sum\limits_{j \in B} \exp{\phi_j}}, i \in B \right).
\end{align}
Additionally, the max \eqref{eq:gumbel_max} and argmax \eqref{eq:gumbel_argmax} are independent. For details, see \citet{maddison2014sampling}.

\subsection{The Gumbel-Top-$k$ trick}
Considering the maximum the \emph{top 1} (one), we can generalize the Gumbel-Max trick to the Gumbel-Top-$k$ trick to draw an \emph{ordered sample of size $k$ without replacement}\footnote{With sampling without replacement from a categorical distribution, we mean sampling the first element, then renormalizing the remaining probabilities to sample the next element, etcetera. This does \emph{not} mean that the \emph{inclusion probability} of element $i$ is proportional to $p_i$: if we sample $k = n$ elements all elements are included with probability 1.}, by taking the indices of the $k$ largest perturbed log-probabilities. 
Generalizing $\argmax$, we denote with $\argtop k$ the function that takes a sequence of values and returns the indices of the $k$ largest values, in order of decreasing value.
\begin{theorem}
\label{thm:gumbel_top_k}
For $k \le n$, let $I^*_1, ..., I^*_k = \argtop k \,{ G_{\phi_i} }$. Then $I^*_1, ..., I^*_k$ is an (ordered) sample without replacement from the $\text{Categorical} \left( \frac{\exp{\phi_i}}{\sum_{j \in N} \exp{\phi_j}}, i \in N \right)$ distribution, e.g. for a realization $i^*_1, ..., i^*_k$ it holds that
\begin{equation}
\label{eq:theorem_topk_trick}
    P\left(I^*_1 = i^*_1, ..., I^*_k = i^*_k\right) = \prod_{j=1}^k \frac{\exp{\phi_{i^*_j}}}{\sum_{\ell \in N^*_j} \exp{\phi_\ell}}
\end{equation}
where $N^*_j = N \setminus \{i^*_1, ..., i^*_{j-1}\}$ is the domain (without replacement) for the $j$-th sampled element.
\end{theorem}

For more info we refer to the blog post by \citet{vieira2014gumbel}. We include a proof in Appendix \ref{app:proof_gumbel_topk_trick}.

\subsection{Sequence models}
A \emph{sequence model} is a factorized parametric distribution over sequences. The parameters $\bm{\theta}$ define the conditional probability $p_{\bm{\theta}}(y_t | \bm{y}_{1:t-1})$ of the next token $y_t$ given the partial sequence $\bm{y}_{1:t-1}$. Typically $p_{\bm{\theta}}(y_t | \bm{y}_{1:t-1})$ is defined as a softmax normalization of unnormalized log-probabilities $\phi_{\bm{\theta}}(y_t | \bm{y}_{1:t-1})$ with optional temperature $T$ (default $T = 1$):
\begin{equation}
\label{eq:softmax}
    p_{\bm{\theta}}(y_t | \bm{y}_{1:l-1}) = \frac{\exp{(\phi_{\bm{\theta}}(y_t | \bm{y}_{1:t-1})/T)}}{\sum_{y'} \exp{(\phi_{\bm{\theta}}(y' | \bm{y}_{1:t-1})/T)}}.
\end{equation}
The normalization is w.r.t.\ a single token, so the model is \emph{locally normalized}. The total probability of a (partial) sequence $\bm{y}_{1:l}$ follows from the chain rule of probability:
\begin{align}
	p_{\bm{\theta}}(\bm{y}_{1:t}) &= p_{\bm{\theta}}(y_t | \bm{y}_{1:t-1}) \cdot p_{\bm{\theta}}(\bm{y}_{1:t-1}) \label{eq:p_model_incremental} \\
	&= \prod\limits_{t'=1}^{t} p_{\bm{\theta}}(y_{t'} | \bm{y}_{1:t'-1}).
	\label{eq:p_model}
\end{align}
A sequence model defines a valid probability distribution over both partial and complete sequences. When the length is irrelevant, we simply write $\bm{y}$ to indicate a (partial or complete) sequence. If the model is additionally conditioned on a \emph{context} $\bm{x}$ (e.g., a source sentence), we write $p_{\bm{\theta}}(\bm{y}|\bm{x})$.

\subsection{Beam Search}
A \emph{beam search} is a limited-width breadth first search. In the context of sequence models, it is often used as an approximation to finding the (single) sequence $\bm{y}$ that maximizes \eqref{eq:p_model}, or as a way to obtain a set of high-probability sequences from the model. Starting from an empty sequence (e.g. $t = 0$), a beam search expands at every step $t = 0, 1, 2, ...$ at most $k$ partial sequences (those with highest probability) to compute the probabilities of sequences with length $t + 1$. It terminates with a beam of $k$ complete sequences, which we assume to be of equal length (as they can be padded).

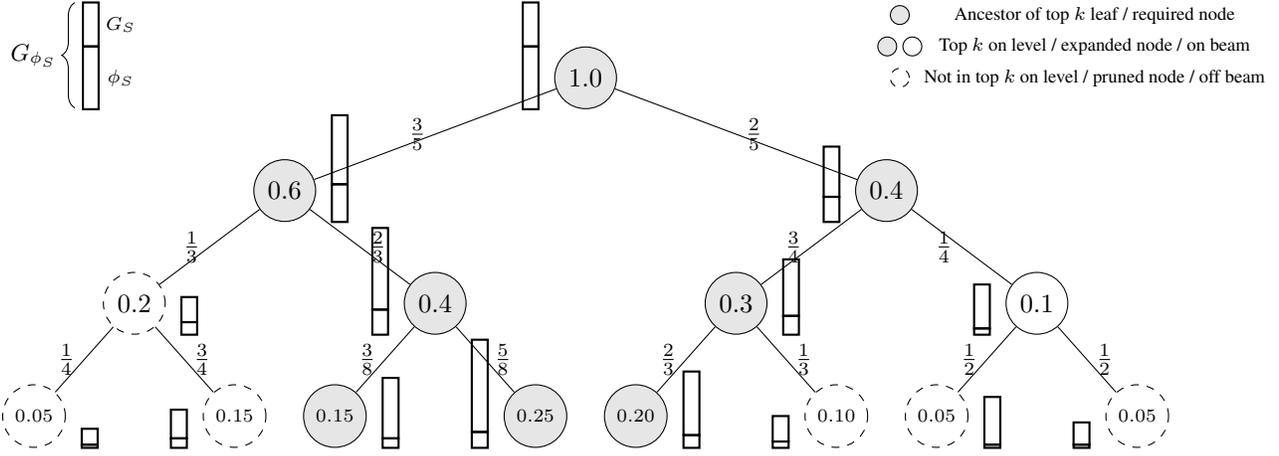
\begin{figure*}[t]
\centering
\begin{tikzpicture}[level/.style={sibling distance=80mm/#1}]
\newcommand{\leafsize}[1]{{\scriptsize{#1}}}

\newcommand\barscale{2.0}
\newcommand{\drawbar}[4]{{\draw[relative to node=#1,thick] (#4,-1) rectangle ++(0.5,\barscale*#2); \draw[relative to node=#1,thick] (#4,-1+\barscale*#2) rectangle ++(0.5,{\barscale*(#3-#2)});}}

\newcommand{\drawbarleft}[4]{{\drawbar{#1}{#2}{#3}{-2}}}
\newcommand{\drawbarright}[4]{{\drawbar{#1}{#2}{#3}{1.5}}}

\tikzset{cut/.style={circle,draw,dashed}}
\tikzset{constructed/.style={circle,draw}}
\tikzset{required/.style={circle,draw,fill=black!10}}

\tikzset{
    relative to node/.style={
        shift={(#1.center)},
        x={(#1.east)},
        y={(#1.north)},
    }
}
\node [required] (root){$1.0$}
  child {node [required] (n0) {$0.6$} 
    child {node [cut] (n00) {$0.2$}
      child {node [cut] (n000) {\leafsize{$0.05$}} edge from parent node[left,draw=none] {$\frac{1}{4}$}} 
      child {node [cut] (n001) {\leafsize{$0.15$}} edge from parent node[right,draw=none] {$\frac{3}{4}$}} 
      edge from parent node[left,draw=none] {$\frac{1}{3}$}
    }
    child {node [required] (n01) {$0.4$}
      child {node [required] (n010) {\leafsize{$0.15$}} edge from parent node[left,draw=none] {$\frac{3}{8}$}}
      child {node [required] (n011) {\leafsize{$0.25$}} edge from parent node[right,draw=none] {$\frac{5}{8}$}} 
      edge from parent node[right,draw=none] {$\frac{2}{3}$}
    }
    edge from parent node[left,draw=none] {$\frac{3}{5}$}
  }
  child {node [required] (n1) {$0.4$}
    child {node [required] (n10) {$0.3$}
      child {node [required] (n100) {\leafsize{$0.20$}} edge from parent node[left,draw=none] {$\frac{2}{3}$}} 
      child {node [cut] (n101) {\leafsize{$0.10$}} edge from parent node[right,draw=none] {$\frac{1}{3}$}} 
      edge from parent node[left,draw=none] {$\frac{3}{4}$}
    }
    child {node [constructed] (n11) {$0.1$}
      child {node [cut] (n110) {\leafsize{$0.05$}} edge from parent node[left,draw=none] {$\frac{1}{2}$}} 
      child {node [cut] (n111) {\leafsize{$0.05$}} edge from parent node[right,draw=none] {$\frac{1}{2}$}} 
      edge from parent node[left,draw=none] {$\frac{1}{4}$}
    }
    edge from parent node[right,draw=none] {$\frac{2}{5}$}
  }
;
\draw[relative to node=root,required] (10,2) circle (0.3);
\draw[relative to node=root] (16.2,2) node {\scriptsize{Ancestor of top $k$ leaf / required node}};

\draw[relative to node=root,required] (9.6,1) circle (0.3);
\draw[relative to node=root,constructed] (10.4,1) circle (0.3);
\draw[relative to node=root] (16.2,1) node {\scriptsize{Top $k$ on level / expanded node / on beam}};

\draw[relative to node=root,cut] (10,0) circle (0.3);
\draw[relative to node=root] (16.2,0) node {\scriptsize{Not in top $k$ on level / pruned node / off beam}};

\drawbar{root}{1.0}{1.7}{-16};
\draw [decorate,decoration={brace,amplitude=5pt},xshift=6pt,yshift=0pt]
(-7,-0.4) -- (-7,1.0) node [black,midway,xshift=-0.55cm] 
{\footnotesize $G_{\phi_S}$};
\draw[relative to node=root] (-14.8,1.7) node {\scriptsize{$G_S$}};
\draw[relative to node=root] (-14.8,0) node {\scriptsize{$\phi_S$}};

\drawbarleft{root}{1.0}{1.7};
\drawbarright{n0}{0.6}{1.7};
\drawbarright{n00}{0.2}{0.6};
\drawbarright{n000}{0.05}{0.3};
\drawbarleft{n001}{0.15}{0.6};
\drawbarleft{n01}{0.4}{1.7};
\drawbarright{n010}{0.15}{1.1};
\drawbarleft{n011}{0.25}{1.7};
\drawbarleft{n1}{0.4}{1.2};
\drawbarright{n10}{0.3}{1.2};
\drawbarright{n100}{0.2}{1.2};
\drawbarleft{n101}{0.1}{0.5};
\drawbarleft{n11}{0.1}{0.8};
\drawbarright{n110}{0.05}{0.8};
\drawbarleft{n111}{0.05}{0.4};

\end{tikzpicture}
\vskip -0.2cm
\caption{Example of the Gumbel-Top-$k$ trick on a tree, with $k = 3$.  The bars next to the leaves indicate the perturbed log-probabilities $G_{\phi_i}$, while the bars next to internal nodes indicate the maximum perturbed log-probability of the set of leaves $S$ in the subtree rooted at that node: $G_{\phi_S} = \max_{i \in S} G_{\phi_i} \sim \text{Gumbel}(\phi_S)$ with $\phi_S = \log \sum_{i \in S} \exp{\phi_i}$. The bar is split in two to illustrate that $G_{\phi_S} = \phi_S + G_S$. Numbers in the nodes represent $p_{\bm{\theta}}(\bm{y}^S) = \exp{\phi_S} = \sum_{i \in S} \exp{\phi_i}$, the probability of the partial sequence $\bm{y}^S$. Numbers at edges represent the conditional probabilities for the next token. The shaded nodes are ancestors of the top $k$ leaves with highest perturbed log-probability $G_{\phi_i}$. These are the ones we actually need to expand. In each layer, there are at most $k$ such nodes, such that we are guaranteed to construct all top $k$ leaves by expanding at least the top $k$ nodes (ranked on $G_{\phi_S}$) in each level (indicated by a solid border).
}
\label{fig:tree}
\vskip -0.4cm
\end{figure*}
\section{Stochastic Beam Search}
We derive Stochastic Beam Search by starting with the \emph{explicit} application of the Gumbel-Top-$k$ trick to sample $k$ sequences without replacement from a sequence model. This requires instantiating all sequences in the domain to find the $k$ largest perturbed log-probabilities. Then we transition to \emph{top-down} sampling of the perturbed log-probabilities, and we use Stochastic Beam Search to instantiate (only) the sequences with the $k$ largest perturbed log-probabilities. As both methods are equivalent, Stochastic Beam Search \emph{implicitly} applies the Gumbel-Top-$k$ trick and thus yields a sample of $k$ sequences without replacement.

\subsection{The Gumbel-Top-$k$ trick on a tree}
We represent the sequence model \eqref{eq:p_model} as a tree (as in Figure \ref{fig:tree}), where internal nodes at level $t$ represent partial sequences $\bm{y}_{1:t}$, and leaf nodes represent completed sequences. We identify a leaf by its index $i \in N = \{1, ..., n\}$ and write $\bm{y}^{i}$ as the corresponding sequence, with (normalized!) log-probability $\phi_i = \log p_{\bm{\theta}}(\bm{y}^{i})$. To obtain a sample from the distribution \eqref{eq:p_model} without replacement, we should sample from the set of leaf nodes $N$ without replacement, for which we can naively use the Gumbel-Top-$k$ trick (Section \ref{sec:gumbel_max_trick}):
\begin{itemize}
  \setlength\itemsep{\itemmargin}
    \item Compute $\phi_i = \log p_{\bm{\theta}}(\bm{y}^{i})$ for \emph{all sequences} $\bm{y}^{i}, i \in N$. To reuse computations for partial sequences, the complete probability tree is instantiated, as in Figure \ref{fig:tree}.
    \item Sample $G_{\phi_i} \sim \text{Gumbel}(\phi_i)$, so $G_{\phi_i}$ can be seen as the perturbed log-probability of sequence $\bm{y}^i$.
    \item Let $i^*_1, ..., i^*_k = \argtop k \, { G_{\phi_i} }$, then $\bm{y}^{i^*_1}, ..., \bm{y}^{i^*_k}$ is a sample of sequences from \eqref{eq:p_model} without replacement.
\end{itemize}
As instantiating the complete probability tree is computationally prohibitive, we construct an equivalent process that only requires computation linear in the number of samples $k$ and the sequence length.

\paragraph{Perturbed log-probabilities of partial sequences.}
For the naive implementation of the Gumbel-Top-$k$ trick, we  only defined the perturbed log-probabilities $G_{\phi_i}$ for leaf nodes $i$, which correspond to \emph{complete sequences} $\bm{y}^i$. For the Stochastic Beam Search implementation, we also define the perturbed log-probabilities for \emph{internal nodes} corresponding to \emph{partial sequences}. We identify a node (internal or leaf) by the set $S$ of leaves in the corresponding subtree, and we write $\bm{y}^S$ as the corresponding (partial or completed) sequence. Its log-probability $\phi_{S} = \log p_{\bm{\theta}}(\bm{y}^S)$ can be computed incrementally from the parent log-probability using \eqref{eq:p_model_incremental}, and since the model is locally normalized, it holds that
\begin{equation}
    \phi_{S} = \log p_{\bm{\theta}}(\bm{y}^S) = \log \sum\limits_{i \in S} \exp \phi_i.
\end{equation}

Now for each node $S$, we define $G_{\phi_S}$ as the \emph{maximum of the perturbed log-probabilities $G_{\phi_i}$ in the subtree leaves $S$}. By Equation \eqref{eq:gumbel_max}, $G_{\phi_S}$ has a Gumbel distribution with location $\phi_S$ (hence its notation $G_{\phi_S}$):
\begin{equation}
\label{eq:gumbel_partial_sequence}
    G_{\phi_S} = \max_{i \in S} G_{\phi_i}  \sim \text{Gumbel}(\phi_S)
\end{equation}
Since $G_{\phi_S} \sim \text{Gumbel}(\phi_S)$ is a Gumbel perturbation of the log-probability $\phi_S = \log p_{\bm{\theta}}(\bm{y}^S)$, we call it the \emph{perturbed log-probability} of the partial sequence $\bm{y}^S$. We can define the corresponding Gumbel noise $G_S \sim  \text{Gumbel}(0)$, which can be inferred from $G_{\phi_S}$ by the relation $G_{\phi_S} = \phi_S + G_S$.

\paragraph{Bottom-up sampling of perturbed log-probabilities.}
We can recursively compute \eqref{eq:gumbel_partial_sequence}. Write $\text{Children}(S)$ as the as the set of direct children of the node $S$ (so $\text{Children}(S)$ is a partition of the set $S$). Since the maximum \eqref{eq:gumbel_partial_sequence} must be attained in one of the subtrees, it holds that 
\begin{equation}
\label{eq:max_children}
    G_{\phi_S} = \max_{S' \in \text{Children}(S)} G_{\phi_{S'}}.
\end{equation}
If we want to sample $G_{\phi_S}$ for all nodes, we can use the \emph{bottom-up} sampling procedure: sample the leaves $G_{\phi_{\{i\}}} = G_{\phi_i}, i \in N$ and recursively compute $G_{\phi_S}$ using \eqref{eq:max_children}. This is effectively sampling from the degenerate (constant) distribution resulting from conditioning on the children.

\paragraph{Top-down sampling of perturbed log-probabilities.}
The recursive bottom-up sampling procedure can be interpreted as ancestral sampling from a tree-structured graphical model (somewhat like Figure \ref{fig:tree}) with edges directed upwards. Alternatively, we can reverse the graphical model and sample the tree \emph{top-down}, starting with the root and recursively sampling the children conditionally. 

Note that for the root $N$ (since it contains all leaves $N$), it holds that $\phi_N = \log \sum_{i \in N} \exp \phi_i = 0$, so we can let $G_{\phi_N} \sim \text{Gumbel}(0)$\footnote{Or we can simply set (e.g. condition on) $G_{\phi_N} = 0$. This does not affect the result by the independence of $\max$ and $\argmax$.}. Starting with $S = N$, we can recursively sample the children conditionally on the parent variable $G_{\phi_S}$. For $S' \in \text{Children}(S)$ it holds that $\phi_{S'} = \log p_{\bm{\theta}}(\bm{y}^{S'})$ and we can sample $G_{\phi_{S'}} \sim \text{Gumbel}(\phi_{S'})$ \emph{conditionally} on \eqref{eq:max_children}, e.g. with \emph{their maximum equal to $G_{\phi_S}$}.

Sampling a set of Gumbels conditionally on their maximum being equal to a certain value is non-trivial, but can be done by first sampling the $\argmax$ and then sampling the individual Gumbels conditionally on both the $\max$ and $\argmax$. Alternatively, we can let $G_{\phi_{S'}} \sim \text{Gumbel}(\phi_{S'})$ independently and let $Z = \max_{S' \in \text{Children}(S)} G_{\phi_{S'}}$. Then
\begin{equation*}
\tilde{G}_{\phi_{S'}} = - \log( \exp( - G_{\phi_S} ) - \exp ( - Z) + \exp ( - G_{\phi_{S'}}) ) \label{eq:trunc_gumb_transform_explicit_main_text}
\end{equation*}
is a set of Gumbels with a maximum equal to $G_{\phi_S}$. See Appendix \ref{app:sampling_gumbels_conditional} for details and numerically stable implementation. 

If we recursively sample the complete tree top-down, this is equivalent to sampling the complete tree bottom-up, and as a result, for all leaves, it holds that $(G_{\phi_{\{i\}}} = ) \, G_{\phi_i} \sim \text{Gumbel}(\phi_i)$, \emph{independently}. The benefit of using top-down sampling is that if we are interested only in obtaining the top $k$ leaves, we do \emph{not} need to instantiate the complete tree.

\paragraph{Stochastic Beam Search}
The key idea of Stochastic Beam Search is to apply the Gumbel-Top-$k$ trick for a sequence model, without instantiating the entire tree, by using top-down sampling. With top-down sampling, to find the top $k$ leaves, at every level in the tree we can suffice with only \emph{expanding} (instantiating the subtree for) the $k$ nodes with highest perturbed log-probability $G_{\phi_S}$. To see this, first assume that we instantiated the complete tree using top-down sampling and consider the nodes that are ancestors of at least one of the top $k$ leaves (the shaded nodes in Figure \ref{fig:tree}). At every level $t$ of the tree, there will be at most $k$ such nodes (as each of the top $k$ leaves has only one ancestor at level $t$), and these nodes will have higher perturbed log-probabilities $G_{\phi_S}$ than the other nodes at level $t$, which do not contain a top $k$ leaf in the subtree. This means that if we discard all but the $k$ nodes with highest log-probabilities $G_{\phi_S}$, we are guaranteed to include the ancestors of the top $k$ leaves. Formally, the $k$-th highest log-probability of the nodes at level $t$ provides a \emph{lower bound} required to be among the top $k$ leaves, while $G_{\phi_S}$ is an \emph{upper bound} for the set of leaves $S$ such that it can be discarded or \emph{pruned} if it is lower than the lower bound, so if $G_{\phi_S}$ is not among the top $k$.

Thus, when we apply the top-down sampling procedure, at each level we only need to expand the $k$ nodes with the highest perturbed log-probabilities $G_{\phi_S}$ to end up with the top $k$ leaves. By the Gumbel-Top-$k$ trick the result is a sample without replacement from the sequence model. The effective procedure is a \emph{beam search} over the (stochastically) perturbed log-probabilities $G_{\phi_S}$ for partial sequences $\bm{y}^S$, hence the name \emph{Stochastic Beam Search}. As we use $G_{\phi_S}$ to select the top $k$ partial sequences, we can also think of $G_{\phi_S}$ as the \emph{stochastic score} of the partial sequence $\bm{y}^S$. We formalize Stochastic Beam Search in Algorithm \ref{alg:stochastic_beam_search}.

\begin{algorithm}[H]
  \centering
  \scriptsize
  \caption{StochasticBeamSearch($p_{\bm{\theta}}$, $k$)} \label{alg:stochastic_beam_search}
  \begin{algorithmic}[1]
  	  \STATE {\bfseries Input:} one-step probability distribution $p_{\bm{\theta}}$, beam/sample size $k$
  	  \STATE \textnormal{Initialize } \textsc{beam} empty
  	  \STATE add $(\bm{y}^N=\emptyset, \phi_N = 0, G_{\phi_N} = 0)$ to \textsc{beam}
  	  \FOR{$t = 1, \ldots, \text{steps}$}
  	    \STATE \textnormal{Initialize } \textsc{expansions} \textnormal{ empty}
  	    \FOR{$(\bm{y}^S, \phi_{S}, G_{\phi_{S}}) \in \textsc{beam}$}
  	        \STATE $Z \gets - \infty$
  	        \FOR {$S' \in \text{Children}(S)$}
  	            \STATE $\phi_{S'} \gets \phi_{S} + \log p_{\bm{\theta}}(\bm{y}^{S'} | \bm{y}^S)$
  	            \STATE $G_{\phi_{S'}} \sim \text{Gumbel}(\phi_{S'})$
  	            \STATE $Z \gets \max(Z, G_{\phi_{S'}})$
  	        \ENDFOR
  	        \FOR{$S' \in \text{Children}(S)$}
  	            \STATE $\tilde{G}_{\phi_{S'}} \gets - \log( \exp( - G_{\phi_S} ) - \exp ( - Z) + \exp ( - G_{\phi_{S'}}) )$
  	            \STATE add $(\bm{y}^{S'}, \phi_{S'}, \tilde{G}_{\phi_{S'}})$ to \textsc{expansions}
  	        \ENDFOR
  	    \ENDFOR
  	    \STATE \textsc{beam} ${} \gets \textnormal{take } \text{top $k$}$ of \textsc{expansions} according to $\tilde{G}$
  	  \ENDFOR
  	  \STATE Return \textsc{beam}
  \end{algorithmic}
\end{algorithm}

\subsection{Relation to Beam Search}
Stochastic Beam Search should not only be considered as a sampling procedure, but also as a principled way to randomize a beam search. As a naive alternative, one could run an ordinary beam search, replacing the top-$k$ operation by sampling. In this scenario, at each step $t$ of the beam search we could sample without replacement from the partial sequence probabilities $p_{\bm{\theta}}(\bm{y}_{1:t})$ using the Gumbel-Top-$k$ trick. 

However, in this naive approach, for a low-probability partial sequence to be extended to completion, it does not only need to be initially chosen, but it will need to be re-chosen, independently, again with low probability, at each step of the beam search. The result is a much lower probability to sample this sequence than assigned by the model. Intuitively, we should somehow \emph{commit} to a sampling `decision' made at step $t$. However, a hard commitment to generate exactly one descendant for each of the $k$ partial sequences at step $t$ would prevent generating any two sequences that share an initial partial sequence.

Our Stochastic Beam Search algorithm makes a \emph{soft} commitment to a partial sequence (node in the tree) by propagating the Gumbel perturbation of the log-probability consistently down the subtree. The partial sequence will then be extended as long as its total perturbed log-probability is among the top $k$, but will fall off the beam if, despite the consistent perturbation, another sequence is more promising. 

\subsection{Relation to rejection sampling}
As an alternative to Stochastic Beam Search, we can draw samples \emph{without} replacement by rejecting duplicates from samples drawn \emph{with} replacement. However, if the domain is large and the entropy low (e.g. if there are only a few valid translations), then rejection sampling requires many samples and consequently many (expensive) model evaluations. Also, we have to estimate how many samples to draw (in parallel) or draw samples sequentially. Stochastic Beam Search executes in a single pass, and requires computation linear in the sample size $k$ and the sequence length, which (except for the beam search overhead) is equal to the computational requirement for sampling with replacement.

\section{Experiments}
\label{sec:experiments}

\subsection{Diverse Beam Search}
\label{seq:diverse_beam_search}
\begin{figure*}[ht]
    \centering
    \begin{subfigure}[b]{0.3\textwidth}
        \includegraphics[width=\textwidth]{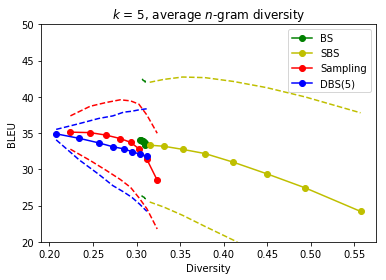}
        \caption{$k = 5$}
        \label{fig:diversity_5}
    \end{subfigure}
    ~
    \begin{subfigure}[b]{0.3\textwidth}
        \includegraphics[width=\textwidth]{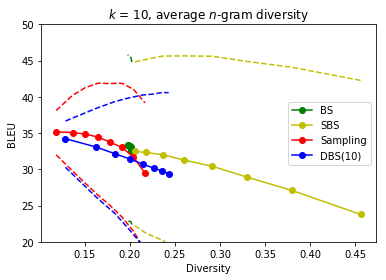}
        \caption{$k = 10$}
        \label{fig:diversity_10}
    \end{subfigure}
    ~
    \begin{subfigure}[b]{0.3\textwidth}
        \includegraphics[width=\textwidth]{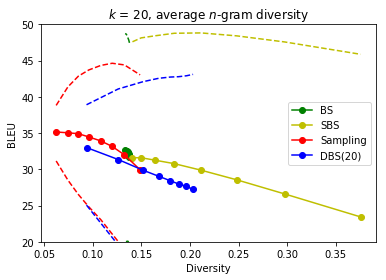}
        \caption{$k = 20$}
        \label{fig:diversity_20}
    \end{subfigure}
\vskip -0.2cm
    \caption{Minimum, mean and maximum BLEU score vs. diversity for different sample sizes $k$. Points indicate different temperatures/diversity strengths, from 0.1 (low diversity, left in graph) to 0.8 (high diversity, right in graph).}\label{fig:diversity}
\vskip -0.4cm
\end{figure*}
In this experiment we compare Stochastic Beam Search as a principled (stochastic) alternative to \emph{Diverse Beam Search} \citep{vijayakumar2018diverse} in the context of neural machine translation to obtain a diverse set of translations for a single source sentence $\bm{x}$. Following the setup by \citet{vijayakumar2018diverse} we report both diversity as measured by the fraction of unique $n$-grams in the $k$ translations as well as mean and maximum BLEU score \citep{papineni2002bleu}  as an indication of the quality of the sample. The maximum BLEU score corresponds to `oracle performance' reported by
\citet{vijayakumar2018diverse}, but we report the mean as well since a single good translation and $k-1$ completely random sentences scores high on both maximum BLEU score and diversity, while being undesirable. A good method should increase diversity without sacrificing mean BLEU score. 

We compare four different sentence generations methods: \emph{Beam Search} (BS), \emph{Sampling}, \emph{Stochastic Beam Search} (SBS) (sampling without replacement) and \emph{Diverse Beam Search} with $G$ groups (DBS($G$)) \citep{vijayakumar2018diverse}. 
For Sampling and Stochastic Beam Search, we control the diversity using the softmax temperature $T$ in Equation \eqref{eq:softmax}. We use $T = 0.1, 0.2, ..., 0.8$, where a higher $T$ results in higher diversity. Heuristically, we also vary $T$ for computing the scores with (deterministic) Beam Search. The diversity of Diverse Beam Search is controlled by the \emph{diversity strengths} parameter, which we vary between $0.1, 0.2, ..., 0.8$. We set the number of groups $G$ equal to the sample size $k$, which \citet{vijayakumar2018diverse} reported as the best choice.

We modify the Beam Search in \path{fairseq} \citep{ott2019fairseq} to implement Stochastic Beam Search\footnote{Our code is available at \url{https://github.com/wouterkool/stochastic-beam-search}}, and use the \path{fairseq} implementations for Beam Search, Sampling and Diverse Beam Search. For theoretical correctness of the Stochastic Beam Search, we disable length-normalization \citep{wu2016google} and early stopping (and therefore also do not use these parameters for the other methods). We use the pretrained model from \citet{gehring2017convolutional} and use the \path{wmt14.v2.en-fr.newstest2014} \footnote{\url{https://s3.amazonaws.com/fairseq-py/data/wmt14.v2.en-fr.newstest2014.tar.bz2}} test set consisting of 3003 sentences. We run the four methods with sample sizes $k=5, 10, 20$ and plot the minimum, mean and maximum BLEU score among the $k$ translations (averaged over the test set) against the average $d = \frac{1}{4} \sum_{n=1}^4 d_n$ of $1,2,3$ and $4$-gram diversity, where $n$-gram diversity is defined as:
\begin{equation*}
    d_n = \frac{\text{\# of unique $n$-grams in $k$ translations}}{\text{total \# of $n$-grams in $k$ translations}}.
\end{equation*}
In Figure \ref{fig:diversity}, we represent the results as curves, indicating the trade-off between diversity and BLEU score. The points indicate datapoints and the dashed lines indicate the (averaged) minimum and maximum BLEU score. For the same diversity, Stochastic Beam Search achieves higher mean/maximum BLEU score. Looking at a certain BLEU score, we observe that Stochastic Beam Search achieves the same BLEU score as Diverse Beam Search with a significantly larger diversity. For low temperatures ($< 0.5$), the maximum BLEU score of Stochastic Beam Search is comparable to the deterministic Beam Search, so the increased diversity does not sacrifice the best element in the sample. Note that Sampling achieves higher mean BLEU score at the cost of diversity, which may be because good translations are sampled repeatedly. However, the maximum BLEU score of both Sampling and Diverse Beam Search is lower than with Beam Search and Stochastic Beam Search.

\subsection{BLEU score estimation}
\begin{figure*}[ht]
    \centering
    \includegraphics[width=\textwidth]{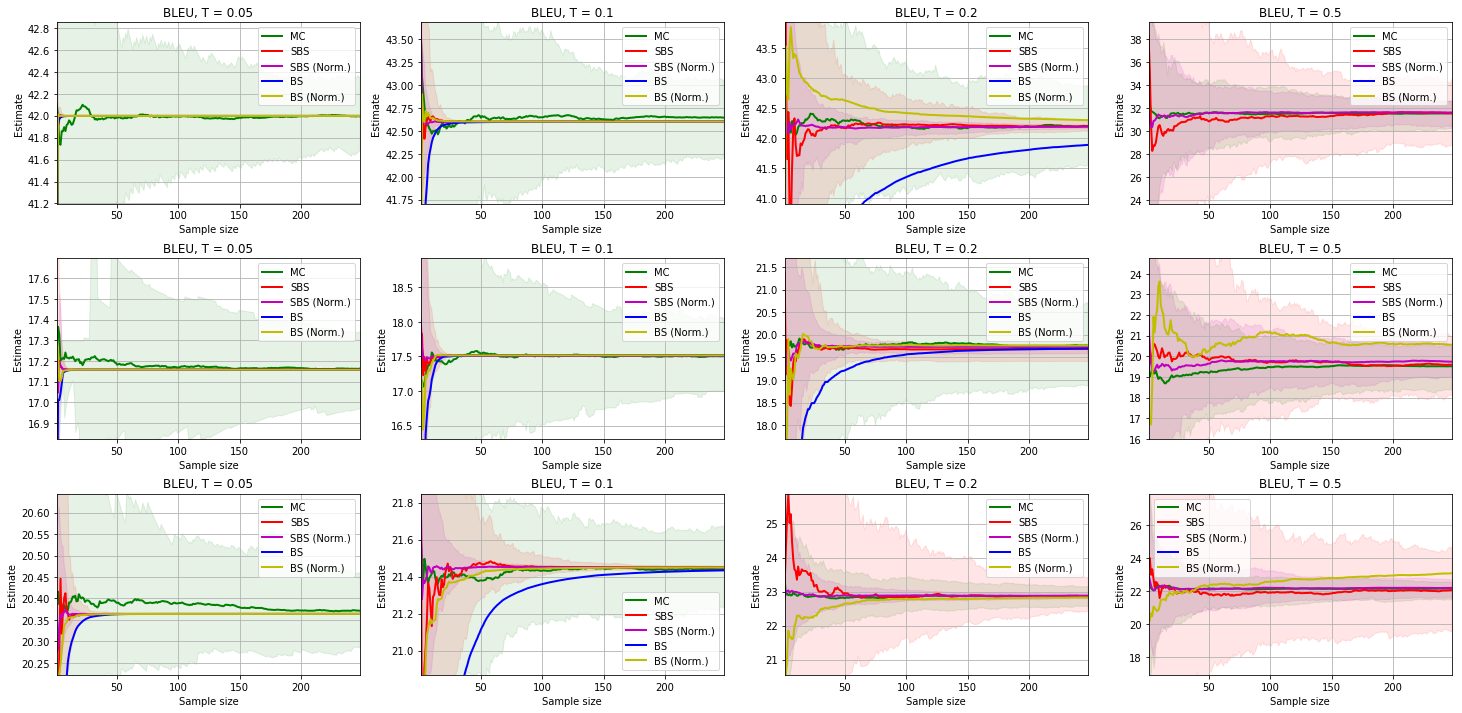}
\vskip -0.2cm
    \caption{BLEU score estimates for three sentences sampled/decoded by different estimators for different temperatures.}
    \label{fig:bleu_estimates}
\vskip -0.4cm
\end{figure*}
In our second experiment, we use sampling without replacement to evaluate the expected \emph{sentence level} BLEU score for a translation $\bm{y}$ given a source sentence $\bm{x}$. Although we are often interested in \emph{corpus level} BLEU score, estimation of sentence level BLEU score is useful, for example when training using minibatches to directly optimize BLEU score \citep{ranzato2016sequence}.

We leave dependence of the BLEU score on the source sentence $\bm{x}$ implicit, and write $f(\bm{y}) = \text{BLEU}(\bm{y},\bm{x})$. Writing the domain of $\bm{y}$ (given $\bm{x}$) as $\bm{y}^1, ..., \bm{y}^n$ (e.g. all possible translations), we want to estimate the following expectation:
\begin{equation}
\label{eq:expectation}
    \mathbb{E}_{\bm{y} \sim p_{\bm{\theta}}(\bm{y}|\bm{x})} \left[f(\bm{y})\right] = \sum\limits_{i=1}^n p_{\bm{\theta}}(\bm{y}^i|\bm{x}) f(\bm{y}^i).
\end{equation}

Under a Monte Carlo (MC) sampling \emph{with replacement} scheme with size $k$, we write $S$ as the set\footnote{Formally, when sampling with replacement, $S$ is a \emph{multiset}.} of indices of sampled sequences $\{\bm{y}^i, i \in S\}$ and estimate \eqref{eq:expectation} using \begin{equation}
\label{eq:estimate_mc}
    \mathbb{E}_{\bm{y} \sim p_{\bm{\theta}}(\bm{y}|\bm{x})} \left[f(\bm{y})\right] \approx \frac{1}{k} \sum_{i \in S} f(\bm{y}^i).
\end{equation}
If the distribution $p_{\bm{\theta}}$ has low entropy (for example if there are only few valid translations), then MC estimation may be inefficient since repeated samples are uninformative. We can use sampling without replacement as an improvement, but we need to use importance weights to correct for the changed sampling probabilities. Using the Gumbel-Top-$k$ trick, we can implement an estimator equivalent to the estimator described by \citet{vieira2017estimating}, derived from priority sampling \citep{duffield2007priority}:
\begin{equation}
\label{eq:BLEU_score_without_replacement}
    \mathbb{E}_{\bm{y} \sim p_{\bm{\theta}}(\bm{y}|\bm{x})} \left[f(\bm{y})\right] \approx \sum_{i \in S}  \frac{p_{\bm{\theta}}(\bm{y}^i|\bm{x})}{q_{\bm{\theta},\kappa}(\bm{y}^i|\bm{x})}f(\bm{y}^i)
\end{equation}
Here $\kappa$ is the $(k+1)$-th largest element of $\{G_{\phi_i}, i \in N\}$, which can be considered the \emph{empirical threshold} for the Gumbel-Top-$k$ trick (since $i \in S$ if $G_{\phi_i} > \kappa$), and we define $q_{\bm{\theta},a}(\bm{y}^i|\bm{x}) = P(G_{\phi_i} > a) = 1 - \exp (-\exp(\phi_i - a))$.
If we would assume a fixed threshold $a$ and \emph{variably} sized sample $S = \{i \in N: G_{\phi_i} > a\}$, then $q_{\bm{\theta},a}(\bm{y}^i|\bm{x}) = P(i \in S)$ and $\frac{p_{\bm{\theta}}(\bm{y}^i|\bm{x})}{q_{\bm{\theta},a}(\bm{y}^i|\bm{x})}$ is a standard importance weight.
Surprisingly, using a \emph{fixed} sample size $k$ (and empirical threshold $\kappa$) also yields in an unbiased estimator, and we include a proof adapted from \citet{duffield2007priority} and \citet{vieira2017estimating} in Appendix \ref{app:unbiased_priority_sampling}. To obtain $\kappa$, we need to sacrifice the last sample\footnote{For Stochastic Beam Search, we use a sample/beam size $k$, set $\kappa$ equal to the $k$-th largest perturbed log-probability and compute \eqref{eq:BLEU_score_without_replacement} based on the remaining $k-1$ samples. Alternatively, we could use a beam size of $k + 1$.}, slightly increasing variance.

Empirically, the estimator \eqref{eq:BLEU_score_without_replacement} has high variance, and in practice it is preferred to normalize the importance weights by $W(S) = \sum_{i \in S}  \frac{p_{\bm{\theta}}(\bm{y}^i|\bm{x})}{q_{\bm{\theta},\kappa}(\bm{y}^i|\bm{x})}$ \citep{hesterberg1988advances}:
\begin{equation}
\label{eq:BLEU_score_without_replacement_normalized}
    \mathbb{E}_{\bm{y} \sim p_{\bm{\theta}}(\bm{y}|\bm{x})} \left[f(\bm{y})\right] \approx \frac{1}{W(S)} \sum_{i \in S}  \frac{p_{\bm{\theta}}(\bm{y}^i|\bm{x})}{q_{\bm{\theta},\kappa}(\bm{y}^i|\bm{x})}f(\bm{y}^i).
\end{equation}
The estimator \eqref{eq:BLEU_score_without_replacement_normalized} is biased but consistent: in the limit $k = n$ we sample the entire domain, so we have empirical threshold $\kappa = -\infty$ and $q_{\bm{\theta},\kappa}(\bm{y}^i|\bm{x}) = 1$ and $W(S) = 1$, such that \eqref{eq:BLEU_score_without_replacement_normalized} is equal to \eqref{eq:expectation}.

We have to take care computing the importance weights as depending on the entropy the terms in the quotient $\frac{p_{\bm{\theta}}(\bm{y}^i|\bm{x})}{q_{\bm{\theta},\kappa}(\bm{y}^i|\bm{x})}$ can become very small, and in our case the computation of $P(G_{\phi_i} > a) = 1 - \exp (-\exp(\phi_i - a))$ can suffer from catastrophic cancellation. For details, see Appendix \ref{app:importance_weights_numeric_stability}.

Because the model is not trained to use its own predictions as input, at test time errors can accumulate. As a result, when sampling with the default temperature $T = 1$, the expected BLEU score is very low (below 10).
To improve quality of generated sentences we use lower temperatures and experiment with  $T = 0.05, 0.1, 0.2, 0.5$. We then use different methods to estimate the BLEU score:
\begin{itemize}
  \setlength\itemsep{\itemmargin}
    \item \emph{Monte Carlo} (MC), using Equation \eqref{eq:estimate_mc}.
    \item \emph{Stochastic Beam Search} (SBS), where we compute estimates using the estimator in Equation \eqref{eq:BLEU_score_without_replacement} and the normalized variant in Equation \eqref{eq:BLEU_score_without_replacement_normalized}.
    \item \emph{Beam Search} (BS), where we compute a deterministic beam $S$ (the temperature $T$ affects the scoring) and compute $\sum_{i \in S} p_{\bm{\theta}}(\bm{y}^i|\bm{x}) f(\bm{y}^i)$. This is not a statistical estimator, but a lower bound to the target \eqref{eq:expectation} which serves as a validation of the implementation and gives insight on how many sequences we need at least to capture most of the mass in \eqref{eq:expectation}. We also compute the normalized version $\frac{\sum_{i \in S} p_{\bm{\theta}}(\bm{y}^i|\bm{x}) f(\bm{y}^i)}{\sum_{i \in S} p_{\bm{\theta}}(\bm{y}^i|\bm{x})}$, which can heuristically be considered as a `determinstic estimate'.
\end{itemize}
In Figure \ref{fig:bleu_estimates} we show the results of computing each estimate 100 times (BS only once as it is deterministic) for three different sentences\footnote{Sentence 1500, 2000 and 2500 from the WMT14 dataset. Sentences 0, 500, 1000 are shorter and obtained 0 BLEU score in some cases, therefore being uninteresting in comparisons.} for temperatures $T = 0.05, 0.1, 0.2, 0.5$ and sample sizes $k = 1$ to $250$. We report the empirical mean and $2.5$-th and $97.5$-th percentile.
The normalized SBS estimator indeed achieves significantly lower variance than the unnormalized version and for $T < 0.5$, it significantly reduces variance compared to MC, without adding observable bias. For $T = 0.5$ the results are similar, but we are less interested in this case as the overall BLEU score is lower than for $T = 0.2$.
\todo{add complete sentences}
\begin{figure*}[ht]
    \centering
    \includegraphics[width=\textwidth]{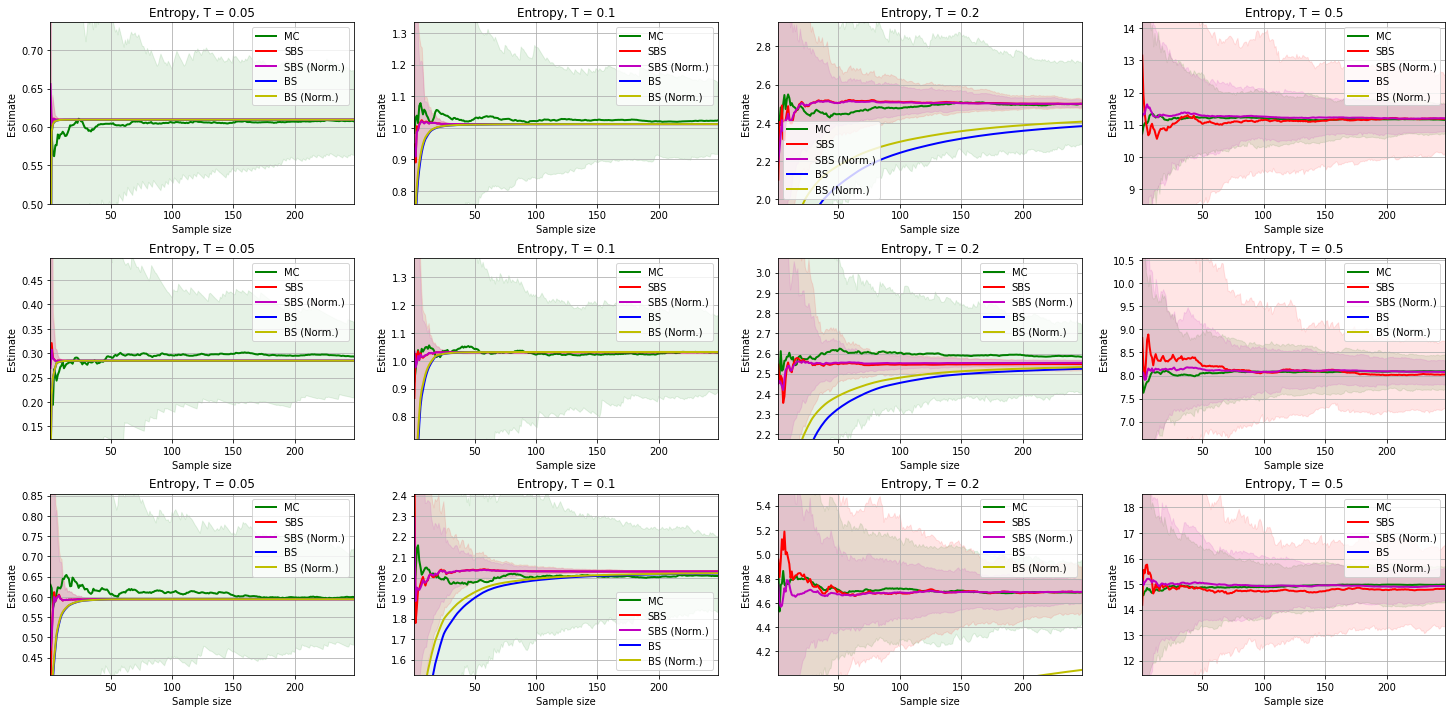}
\vskip -0.2cm
    \caption{Entropy score estimates for three sentences sampled/decoded by different estimators for different temperatures.}
    \label{fig:entropy}
\vskip -0.4cm
\end{figure*}

\subsection{Conditional Entropy Estimation}
Additionally to estimating the BLEU score we use $f(\bm{y}) = -\log p_{\bm{\theta}}(\bm{y}|\bm{x})$ such that Equation \eqref{eq:expectation} becomes the model entropy (conditioned on the source sentence $\bm{x}$):
\begin{equation*}
    \mathbb{E}_{\bm{y} \sim p_{\bm{\theta}}(\bm{y}|\bm{x})} \left[-\log p_{\bm{\theta}}(\bm{y}|\bm{x})\right].
\end{equation*}
Entropy estimation is useful in optimization settings where we want to include an entropy loss to ensure diversity. It is a different problem than BLEU score estimation as high BLEU score (for a good model) correlates positively with model probability, while for entropy rare $\bm{y}$ contribute the largest terms $-\log p_{\bm{\theta}}(\bm{y}|\bm{x})$. We use the same experimental setup as for the BLEU score and present the results in Figure \ref{fig:entropy}. The results are similar to the BLEU score experiment: the normalized SBS estimate has significantly lower variance than MC for $T < 0.5$ while for $T = 0.5$, results are similar. This shows that Stochastic Beam Search can be used to construct practical statistical estimators.

\section{Related Work}

\subsection{Sampling and the Gumbel-Max trick}
The idea of sampling by solving optimization problems has been used for various purposes \citep{papandreou2011perturb, hazan2012partition, tarlow2012randomized, ermon2013embed,  maddison2014sampling, chen2016scalable, balog2017lost}, but to our knowledge this idea has not been used for sampling without replacement.

Most closely related to our work, \citet{maddison2014sampling} note that the Gumbel-Max trick \citep{gumbel1954statistical} can be applied implicitly and generalize it to continuous distributions, using an $A^*$ search to find the maximum of a \emph{Gumbel} process.
In this work, we extend the idea of top-down sampling to efficiently draw \emph{multiple samples without replacement} from a  \emph{factorized distribution} (with possibly exponentially large domain) by implicitly applying the Gumbel-Top-$k$ trick.
This is a new and practical sampling method.

The blog post by \citet{vieira2014gumbel} describes the relation of the Gumbel-Top-$k$ trick (as we call it) to Weighted Reservoir Sampling \citep{efraimidis2006weighted},  an algorithm for drawing weighted samples without replacement, either from a stream or efficiently in parallel. When sampling the complete domain, this is equivalent to the Thurstonian \citep{thurstone1927} interpretation of the Plackett-Luce ranking model \citep{plackett1975analysis,luce1959individual} as given by \citet{yellott1977relationship}. The connection of the Plackett-Luce model to the Gumbel-Max trick and the implication that this can be used for sampling without replacement is not widely known\footnote{For example, currently the popular PyTorch \citep{paszke2017automatic} library uses the (for large $k$) expensive sequential algorithm.}.

The Gumbel-Max trick has also been used to define relaxations of the categorical distribution \citep{maddison2016concrete, jang2016categorical}, which can be reparameterized for low-variance but biased gradient estimators. Recently, \citet{grover2018stochastic} extended this idea to the Plackett-Luce distribution to optimize stochastic sorting networks. We think our work is a step in the direction to improve these methods in the context of sequence models \citep{gu2017neural}.
\todoherke{Think I would try to end with how it relates to your work. Either explictly comparing, or just saying what the shortcomings are (which your method doesn't have) }

\subsection{Beam search}
Beam search is widely used for approximate inference in various domains such as machine translation \citep{sutskever2014sequence,bahdanau2014neural,ranzato2016sequence,vaswani2017attention,gehring2017convolutional}, image captioning \citep{vinyals2015show}, speech recognition \citep{graves2013speech} and other structured prediction settings \citep{vinyals2015pointer,weiss2015structured}. Although typically a test-time procedure, there are works that include beam search in the training loop \citep{daume2009search,wiseman2016sequence,edunov2018classical,negrinho2018learning,edunov2018understanding} for training sequence models on the sequence level \citep{ranzato2016sequence,bahdanau2017actor}.
Many variants of beam search have been developed, such as a continuous relaxation \citep{goyal18continuous}, diversity encouraging variants \citep{li2016simple,shao2017generating,vijayakumar2018diverse} or using modifications such as length-normalization \citep{wu2016google} or simply applying noise to the output \citep{edunov2018understanding}.
Our Stochastic Beam Search is a principled alternative that shares some of the benefits of these heuristic variants, such as the ability to control diversity or produce randomized output.
\todoherke{Think I would try to end with how it relates to your work. Either explictly comparing, or just saying what the shortcomings are (which your method doesn't have) }

\section{Discussion}
\todoherke{Reformulated, original in comments if you want to revert easily..}
We introduced \emph{Stochastic Beam Search}: an algorithm that is easy to implement on top of a beam search as a way to sample sequences without replacement. This algorithm relates sampling and beam search, combining advantages of these two methods. Our experiments support the idea that it can be used as a drop-in replacement in places where sampling or beam search is used. In fact, our experiments show Stochastic Beam Search can be used to yield lower-variance estimators and high-diversity samples from a neural machine translation model. In future work, we plan to leverage the probabilistic interpretation of beam search to develop new beam search related statistical learning methods.  

\bibliography{references}
\bibliographystyle{icml2019}

%% file: appendix.tex
\appendix
\section{Proof of the Gumbel-Top-$k$ trick}
\label{app:proof_gumbel_topk_trick}

\begingroup
\def\thetheorem{\ref{thm:gumbel_top_k}}
\begin{theorem}
For $k \le n$, let $I^*_1, ..., I^*_k = \argtop k \,{ G_{\phi_i} }$. Then $I^*_1, ..., I^*_k$ is an (ordered) sample without replacement from the $\text{Categorical} \left( \frac{\exp{\phi_i}}{\sum_{j \in N} \exp{\phi_j}}, i \in N \right)$ distribution, e.g. for a realization $i^*_1, ..., i^*_k$ it holds that
\begin{equation}
\label{eq:theorem_topk_trick_app}
    P\left(I^*_1 = i^*_1, ..., I^*_k = i^*_k\right) = \prod_{j=1}^k \frac{\exp{\phi_{i^*_j}}}{\sum_{\ell \in N^*_j} \exp{\phi_\ell}}
\end{equation}
where $N^*_j = N \setminus \{i^*_1, ..., i^*_{j-1}\}$ is the domain (without replacement) for the $j$-th sampled element.
\end{theorem}
\addtocounter{theorem}{-1}
\endgroup

\begin{proof}
\todoherke{repeat the theorem here?}First note that
\begin{align}
    & \, P\left(I^*_k = i^*_k \middle| I^*_1 = i^*_1, ..., I^*_{k-1} = i^*_{k-1} \right) \nonumber \\
    =& \, P\left(i^*_k = \argmax_{i \in N^*_k}{ G_{\phi_i} } \middle| I^*_1 = i^*_1, ..., I^*_{k-1} = i^*_{k-1} \right) \nonumber \\
    =& \, P\left(i^*_k = \argmax_{i \in N^*_k}{ G_{\phi_i} } \middle|  \max_{i \in N^*_k}{ G_{\phi_i} } < G_{\phi_{i^*_{k-1}}}\right) \label{eq:topk_proof_before_independence} \\
    =& \, P\left(i^*_k = \argmax_{i \in N^*_k}{ G_{\phi_i} }\right) \label{eq:topk_proof_independence} \\
    =& \, \frac{\exp{\phi_{i^*_k}}}{\sum_{\ell \in N^*_k} \exp{\phi_\ell}} \label{eq:proof_gumbel_max_trick}. 
\end{align}
The step from \eqref{eq:topk_proof_before_independence} to \eqref{eq:topk_proof_independence} follows from the independence of the $\max$ and $\argmax$ (Section \ref{sec:gumbel_max_trick}) and the step from \eqref{eq:topk_proof_independence} to \eqref{eq:proof_gumbel_max_trick} uses the Gumbel-Max trick.
The proof follows by induction on $k$. The case $k = 1$ is the Gumbel-Max trick, while if we assume the result \eqref{eq:theorem_topk_trick_app} proven for $k - 1$, then
\begin{align}
    & \, P\left(I^*_1 = i^*_1, ..., I^*_k = i^*_k\right) \nonumber\\
    = & \, P\left(I^*_k = i^*_k \middle| I^*_1 = i^*_1, ..., I^*_{k-1} = i^*_{k-1}\right) \nonumber\\
    & \cdot P\left(I^*_1 = i^*_1, ..., I^*_{k-1} = i^*_{k-1}\right) \nonumber\\
    = & \, \frac{\exp{\phi_{i^*_k}}}{\sum_{\ell \in N^*_k} \exp{\phi_\ell}} \cdot \prod_{j=1}^{k-1} \frac{\exp{\phi_{i^*_j}}}{\sum_{\ell \in N^*_j} \exp{\phi_\ell}} 
    \label{eq:topk_induction_gumbel_max} \\
    = & \, \prod_{j=1}^k \frac{\exp{\phi_{i^*_j}}}{\sum_{\ell \in N^*_j} \exp{\phi_\ell}}. \nonumber
\end{align}
In \eqref{eq:topk_induction_gumbel_max} we have used Equation \eqref{eq:proof_gumbel_max_trick} and Equation \eqref{eq:theorem_topk_trick_app} for $k - 1$ by induction.
\end{proof}

\newpage

\section{Sampling set of Gumbels with maximum $T$}
\label{app:sampling_gumbels_conditional}

\subsection{The truncated Gumbel distribution}
A random variable $G'$ has a \emph{truncated} Gumbel distribution with location $\phi$ and maximum $T$ (e.g. $G' \sim \text{TruncatedGumbel}(\phi, T)$) with CDF $F_{\phi,T}(g)$ if:
\begin{align}
    & \,F_{\phi,T}(g) \nonumber \\
    =& \, P(G' \le g) \nonumber \\
    =& \, P(G \le g | G \le T) \nonumber \\
    =& \, \frac{P(G \le g \cap G \le T)}{P(G \le T)} \nonumber\\
    =& \, \frac{P(G \le \min(g, T))}{P(G \le T)} \nonumber \\
    =& \, \frac{F_\phi(\min(g, T))}{F_\phi(T)} \nonumber\\
    =& \, \frac{\exp (- \exp (\phi - \min(g, T)))}{\exp (- \exp (\phi - T))} \nonumber \\
    =& \, \exp (\exp (\phi - T) - \exp (\phi - \min(g, T))). \label{eq:trunc_gumb_cdf}
\end{align}
The inverse CDF is:
\begin{equation}
\label{eq:trunc_gumb_inv_cdf}
    F^{-1}_{\phi,T}(u) = \phi - \log( \exp( \phi - T ) - \log u ).
\end{equation}

\subsection{Sampling set of Gumbels with maximum $T$}
In order to sample a set of Gumbel variables $\{\tilde{G}_{\phi_i}|\max_i \tilde{G}_{\phi_i} = T\}$, e.g. with their maximum being \emph{exactly} $T$, we can first sample the $\argmax$, $i^*$ and then sample the Gumbels conditionally on both the $\max$ and $\argmax$:
\begin{enumerate}
    \item Sample $i^* \sim \text{Categorical} \left( \frac{\exp{\phi_i}}{\sum_{j} \exp{\phi_j}} \right)$. We do not need to condition on $T$ since the $\argmax$ $i^*$ is independent of the $\max$ $T$ (Section \ref{sec:gumbel_max_trick}).
    \item Set $\tilde{G}_{\phi_{i^*}} = T$, since this follows from conditioning on the $\max$ $T$ and $\argmax$ $i^*$.
    \item Sample $\tilde{G}_{\phi_i} \sim \text{TruncatedGumbel}(\phi_i, T)$ for $i \neq i^*$. This works because, conditioning on the $\max$ $T$ and $\argmax$ $i^*$, it holds that:
\begin{align*}
    &P(\tilde{G}_{\phi_i} < g | \max_i \tilde{G}_{\phi_i} = T, \argmax_i \tilde{G}_{\phi_i} = i^*, i \neq i^*) \\
    &= P(\tilde{G}_{\phi_i} < g | \tilde{G}_{\phi_i} < T).
\end{align*}
\end{enumerate}

Equivalently, we can let $G_{\phi_i} \sim \text{Gumbel}(\phi_i)$, let $Z = \max_i G_{\phi_i}$ and define
\begin{align}
    \tilde{G}_{\phi_i} &= F^{-1}_{\phi_i,T}(F_{\phi_i,Z}(G_{\phi_i})) \nonumber\\
    &= \phi_i - \log( \exp( \phi_i - T ) \nonumber \\
    &\phantom{{}=\phi_i \log (} - \exp (\phi_i - Z) + \exp (\phi_i - G_{\phi_i}) ) \notag \\
    &= - \log( \exp( - T ) - \exp ( - Z) + \exp ( - G_{\phi_i}) ). \label{eq:app_trunc_gumb_transform_explicit}
\end{align}
Here we have used \eqref{eq:trunc_gumb_cdf} and \eqref{eq:trunc_gumb_inv_cdf}. Since the transformation \eqref{eq:app_trunc_gumb_transform_explicit} is monotonically increasing, it preserves the $\argmax$ and it follows from the Gumbel-Max trick \eqref{eq:gumbel_argmax} that 
\begin{equation*}
    \argmax_i \tilde{G}_{\phi_i} = \argmax_i G_{\phi_i} \sim \text{Categorical} \left( \frac{\exp{\phi_i}}{\sum_{j} \exp{\phi_j}} \right).
\end{equation*}
We can think of this as using the Gumbel-Max trick for step 1 (sampling the argmax) in the sampling process described above.
Additionally, for $i = \argmax_i G_{\phi_i}$:
\begin{equation*}
    \tilde{G}_{\phi_{i}} = F^{-1}_{\phi_i,T}(F_{\phi_i,Z}(G_{\phi_i})) = F^{-1}_{\phi_i,T}(F_{\phi_i,Z}(Z)) = T
\end{equation*}
so here we recover step 2 (setting $\tilde{G}_{\phi_{i^*}} = T$). 
For $i \neq \argmax_i G_{\phi_i}$ it holds that:
\begin{align*}
    &P(\tilde{G}_{\phi_i} \le g|i \neq \argmax_i G_{\phi_i}) \\
    =& \mathbb{E}_Z(P(\tilde{G}_{\phi_i} \le g|Z, i \neq \argmax_i G_{\phi_i})) \\
    =& \mathbb{E}_Z(P(\tilde{G}_{\phi_i} \le g|Z, G_{\phi_i} < Z)) \\
    =& \mathbb{E}_Z(P(F^{-1}_{\phi_i,T}(F_{\phi_i,Z}(G_{\phi_i})) \le g|Z, G_{\phi_i} < Z)) \\
    =& \mathbb{E}_Z(P(G_{\phi_i} \le F^{-1}_{\phi_i,Z}(F_{\phi_i,T}(g))|Z, G_{\phi_i} < Z)) \\
    =& \mathbb{E}_Z(F_{\phi_i,Z}(F^{-1}_{\phi_i,Z}(F_{\phi_i,T}(g)))) \\
    =& \mathbb{E}_Z(F_{\phi_i,T}(g)) = F_{\phi_i,T}(g).
\end{align*}
This means that $\tilde{G}_{\phi_i} \sim \text{TruncatedGumbel}(\phi_i, T)$, so this is equivalent to step 3 (sampling $\tilde{G}_{\phi_i} \sim \text{TruncatedGumbel}(\phi_i, T)$ for $i \neq i^*$).

\subsection{Numeric stability of truncated Gumbel computation}
\label{app:trunc_gumbel_numeric_stability}
Direct computation of \eqref{eq:app_trunc_gumb_transform_explicit} can be unstable as large terms need to be exponentiated. Instead, we compute:
\begin{align}
    v_i &= T - G_{\phi_i} + \logonemexp(G_{\phi_i} - Z) \label{eq:sample_truncated_gumbel_vi} \\
    \tilde{G}_{\phi_i} &= T - \max(0, v_i) - \logonepexp(-|v_i|)  \label{eq:sample_truncated_gumbel_stable}
\end{align}
where we have defined
\begin{align*}
    \logonemexp(a) &= \log(1-\exp(a)), \quad a \le 0 \\
    \logonepexp(a) &= \log(1+\exp(a)).
\end{align*}

This is equivalent as
\begin{align*}
    & \, T - \max(0, v_i) - \log(1 + \exp(-|v_i|)) \\
    =& \, T - \log(1 + \exp(v_i)) \\
    =& \, T - \log\left(1 + \exp\left(T - G_{\phi_i} + \log\left(1-\exp\left(G_{\phi_i} - Z\right)\right)\right)\right) \\
    =& \, T - \log\left(1 + \exp\left(T - G_{\phi_i}\right) \left(1-\exp\left(G_{\phi_i} - Z\right)\right)\right) \\
    =& \, T - \log\left(1 + \exp\left(T - G_{\phi_i}\right) - \exp\left(T - Z\right)\right) \\
    =& \, -\log\left(\exp(-T) + \exp(- G_{\phi_i}) - \exp(- Z)\right) \\
    =& \, \tilde{G}_{\phi_i} \\
\end{align*}
The first step can be easily verified by considering the cases $v_i < 0$ and $v_i \ge 0$. 
$\logonemexp$ and $\logonepexp$ can be computed accurately using $\logonep(a) = \log(1+a)$ and $\expmone(a) = \exp(a) - 1$ \citep{machler2012accurately}:
\begin{align*}
    \logonemexp(a) &= \begin{cases}
    \log(-\expmone(a)) & a > -0.693 \\
    \logonep(-\exp(a)) & \text{otherwise} \\
    \end{cases} \\
    \logonepexp(a) &= \begin{cases}
    \logonep(\exp(a)) & a < 18 \\
    x + \exp(a) & \text{otherwise} \\
    \end{cases}
\end{align*}

\section{Numerical stability of importance weights}
\label{app:importance_weights_numeric_stability}
We have to take care computing the importance weights as depending on the entropy the terms in the quotient $\frac{p_{\bm{\theta}}(\bm{y}_i|\bm{x})}{q_{\bm{\theta}}(\bm{y}_i|\bm{x})}$ can become very small, and in our case the computation of $P(G_{\phi_i} > \kappa) = 1 - \exp (-\exp(\phi_i - \kappa))$ can suffer from catastrophic cancellation. We can rewrite this expression using the more numerically stable implementation $\text{exp1m}(x) = \exp(x) - 1$ as $p(G_{\phi_i} > \kappa) = -\text{exp1m}(-\exp(\phi_i - \kappa))$ but in some cases this still suffers from instability as $\exp(\phi_i - \kappa)$ can underflow if $\phi_i - \kappa$ is small. Instead, for $\phi_i - \kappa < -10$ we use the identity
\begin{equation*}
    \log (1 - \exp(-z)) = \log(z) - \frac{z}{2} + \frac{z^2}{24} - \frac{z^4}{2880} + \bigO(z^6)
\end{equation*}
to directly compute the log importance weight using $z = \exp(\phi_i - \kappa)$ and $\phi_i = \log p_{\bm{\theta}}(\bm{y}_i|\bm{x})$ (we assume $\phi_i$ is normalized):
\begin{align*}
    & \, \log\left(\frac{p_{\bm{\theta}}(\bm{y}_i|\bm{x})}{q_{\bm{\theta}}(\bm{y}_i|\bm{x})}\right) = \log p_{\bm{\theta}}(\bm{y}_i|\bm{x}) - \log q_{\bm{\theta}}(\bm{y}_i|\bm{x}) \\
    =& \, \log p_{\bm{\theta}}(\bm{y}_i|\bm{x}) - \log \left( 1 - \exp (-\exp(\phi_i - \kappa)) \right) \\
    =& \, \log p_{\bm{\theta}}(\bm{y}_i|\bm{x}) - \log \left( 1 - \exp (-z) \right) \\
    =& \, \log p_{\bm{\theta}}(\bm{y}_i|\bm{x}) - \left( \log(z) - \frac{z}{2} + \frac{z^2}{24} - \frac{z^4}{2880} + \bigO(z^6) \right) \\
    =& \, \log p_{\bm{\theta}}(\bm{y}_i|\bm{x})  - \left( \phi_i - \kappa - \frac{z}{2} + \frac{z^2}{24} - \frac{z^4}{2880} + \bigO(z^6) \right) \\
    =& \, \kappa + \frac{z}{2} - \frac{z^2}{24} + \frac{z^4}{2880} + \bigO(z^6)
\end{align*}
If $\phi_i - \kappa < -10$ then $0 < z < 10^{-6}$ so this computation will not lose any significant digits.

\clearpage
\section{Proof of unbiasedness of priority sampling estimator}
\label{app:unbiased_priority_sampling}
The following proof is adapted from the proofs by \citet{duffield2007priority} and \citet{vieira2017estimating}. For generality of the proof, we write $f(i) = f(\bm{y}^i)$, $p_i = p_{\bm{\theta}}(\bm{y}^i|\bm{x})$ and $q_i(\kappa) = q_{\bm{\theta},\kappa}(\bm{y}^i|\bm{x})$, and we consider general keys $h_i$ (not necessarily Gumbel perturbations). 

We assume we have a probability distribution over a finite domain $1, ..., n$ with normalized probabilities $p_i$, e.g. $\sum_{i=1}^n p_i = 1$. For a given function $f(i)$ we want to estimate the expectation
\begin{equation*}
    \mathbb{E}[f(i)] = \sum_{i=1}^n p_i f(i).
\end{equation*}

Each element $i$ has an associated random key $h_i$ and we define $q_i(a) = P(h_i > a)$. This way, if we \emph{know} the threshold $a$ it holds that $q_i(a) = P(i \in S)$ is the probability that element $i$ is in the sample $S$. As was noted by \citet{vieira2017estimating}, the actual distribution of the key does not influence the unbiasedness of the estimator but does determine the effective sampling scheme. Using the Gumbel perturbed log-probabilities as keys (e.g. $h_i = G_{\phi_i}$) is equivalent to the PPSWOR scheme described by \citet{vieira2017estimating}.

We define shorthand notation $h_{1:n} = \{h_1, ..., h_n\}$, $h_{-i} = \{h_1, ..., h_{i-1}, h_{i+1}, ..., h_n\} = h_{1:n} \setminus \{h_i\}$. For a given sample size $k$, let $\kappa$ be the $(k+1)$-th largest element of $h_{1:n}$, so $\kappa$ is the \emph{empirical threshold}. Let $\kappa'_i$ be the $k$-th largest element of $h_{-i}$ (the $k$-th largest of all \emph{other} elements).

Similar to \citet{duffield2007priority} we will show that every element $i$ in our sample contributes an unbiased estimate of $\mathbb{E}[f(i)]$, so that the total estimator is unbiased. Formally, we will prove that
\begin{equation}
\label{eq:each_term_unbiased}
    \mathbb{E}_{h_{1:n}}\left[\frac{\mathbbm{1}_{\{i \in S\}}}{q_i(\kappa)}\right] = 1
\end{equation}
from which the result follows:
\begin{align*}
    & \, \mathbb{E}_{h_{1:n}} \left[ \sum_{i \in S} \frac{p_i}{q_i(\kappa)} f(i) \right] \\
    = & \, \mathbb{E}_{h_{1:n}} \left[ \sum_{i=1}^n \frac{p_i}{q_i(\kappa)} f(i) \mathbbm{1}_{\{i \in S\}} \right] \\
    = & \, \sum_{i=1}^n p_i f(i) \cdot \mathbb{E}_{h_{1:n}} \left[ \frac{\mathbbm{1}_{\{i \in S\}}}{q_i(\kappa)} \right] \\
    = & \, \sum_{i=1}^n p_i f(i) \cdot 1 = \sum_{i=1}^n p_i f(i) = \mathbb{E}[f(i)]
\end{align*}
To prove \eqref{eq:each_term_unbiased}, we make use of the observation (slightly rephrased) by \citet{duffield2007priority} that conditioning on $h_{-i}$, we know $\kappa'_i$ and the event $i \in S$ implies that $\kappa = \kappa'_i$ since $i$ will only be in the sample if $h_i > \kappa'_i$ which means that $\kappa'_i$ is the $k+1$-th largest value of $h_{-i} \cup \{h_i\} = h_{1:n}$. The reverse is also true (if $\kappa = \kappa'_i$ then $h_i$ must be larger than $\kappa'_i$ since otherwise the $k+1$-th largest value of $h_{1:n}$ will be smaller than $\kappa'_i$).
\begin{align*}
    & \, \mathbb{E}_{h_{1:n}}\left[\frac{\mathbbm{1}_{\{i \in S\}}}{q_i(\kappa)}\right] \\
    = & \, \mathbb{E}_{h_{-i}}\left[\mathbb{E}_{h_i}\left[\frac{\mathbbm{1}_{\{i \in S\}}}{q_i(\kappa)}\middle|h_i\right]\right] \\
    \begin{split}
    = & \, \mathbb{E}_{h_{-i}}\left[\mathbb{E}_{h_i}\left[\frac{\mathbbm{1}_{\{i \in S\}}}{q_i(\kappa)}\middle|h_{-i},i \in S\right]P(i \in S | h_{-i}) \right. \\
    & \hphantom{\mathbb{E}_{h_{-i}}}\left. \, + \, \mathbb{E}_{h_i}\left[\frac{\mathbbm{1}_{\{i \in S\}}}{q_i(\kappa)}\middle|h_{-i},i \not\in S\right]P(i \not\in S | h_{-i})
    \right]
    \end{split} \\
    = & \, \mathbb{E}_{h_{-i}}\left[\mathbb{E}_{h_i}\left[\frac{1}{q_i(\kappa)}\middle|h_{-i},i \in S\right]P(i \in S | h_{-i}) + 0 \right] \\
    = & \, \mathbb{E}_{h_{-i}}\left[\mathbb{E}_{h_i}\left[\frac{1}{q_i(\kappa)}\middle|h_{-i},i \in S\right] q_i(\kappa'_i) \right] \\
    = & \, \mathbb{E}_{h_{-i}}\left[\mathbb{E}_{h_i}\left[\frac{1}{q_i(\kappa)}\middle|\kappa = \kappa'_i \right] q_i(\kappa'_i) \right] \\
    = & \, \mathbb{E}_{h_{-i}}\left[\mathbb{E}_{h_i}\left[\frac{1}{q_i(\kappa'_i)} \right] q_i(\kappa'_i) \right] \\
    = & \, \mathbb{E}_{h_{-i}}\left[\frac{1}{q_i(\kappa'_i)} q_i(\kappa'_i) \right] = \mathbb{E}_{h_{-i}}\left[ 1 \right] = 1 \\
\end{align*}